\documentclass[lettersize,journal]{IEEEtran}
\usepackage{amsmath,amsfonts}
\usepackage{algorithmic}
\usepackage{array}
\usepackage{textcomp}
\usepackage{stfloats}
\usepackage{url}
\usepackage{verbatim}
\usepackage{graphicx}
\usepackage{pgfplots}
	\pgfplotsset{compat=1.12}
\usepackage{tikz}
	\usetikzlibrary{positioning}
	\usepackage{tikz-3dplot} 
	\usetikzlibrary{shapes,arrows, positioning, automata, shadows,arrows.meta,backgrounds,fit, calc}
	\usepackage{tkz-fct, tikz-dimline}
	\usepackage{subcaption}
\usepackage{multirow}
	
\def\BibTeX{{\rm B\kern-.05em{\sc i\kern-.025em b}\kern-.08em
    T\kern-.1667em\lower.7ex\hbox{E}\kern-.125emX}}
\usepackage{balance}
\usepackage{soul} 
\usepackage{setspace}

\newcolumntype{C}[1]{>{\centering\let\newline\\\arraybackslash\hspace{0pt}}m{#1}}

\begin{document}
\title{Joint Geometry and Attribute Upsampling of Point Clouds Using Frequency-Selective Models with Overlapped Support}
\author{Viktoria Heimann, Andreas Spruck, and Andr\'e Kaup \vspace{-.2cm}
\thanks{Manuscript created 14 October 2022.\\
	The authors are with the Chair of Multimedia Communications and Signal Processing, Friedrich-Alexander Universit\"at, Erlangen-N\"urnberg (FAU), 91058 Erlangen, Germany (e-mail: viktoria.heimann@fau.de; andreas.spruck@fau.de; andre.kaup@fau.de).\\
	This work was partly funded by the Deutsche Forschungsgemeinschaft (DFG, German Research Foundation) -- SFB 1483 -- Project-ID 442419336, EmpkinS.
	}}


\maketitle

\begin{abstract}
	With the increasing demand of capturing our environment in three-dimensions for AR/ VR applications and autonomous driving among others, the importance of high-resolution point clouds rises. As the capturing process is a complex task, point cloud upsampling is often desired. We propose Frequency-Selective Upsampling (FSU), an upsampling scheme that upsamples geometry and attribute information of point clouds jointly in a sequential manner with overlapped support areas. The point cloud is partitioned into blocks with overlapping support area first. Then, a continuous frequency model is generated that estimates the point cloud's surface locally. The model is sampled at new positions for upsampling. In a subsequent step, another frequency model is created that models the attribute signal. Here, knowledge from the geometry upsampling is exploited for a simplified projection of the points in two dimensions. The attribute model is evaluated for the upsampled geometry positions. In our extensive evaluation, we evaluate geometry and attribute upsampling independently and show joint results. The geometry results show best performances for our proposed FSU in terms of point-to-plane error and plane-to-plane angular similarity. Moreover, FSU outperforms other color upsampling schemes by \(1.9\)~dB in terms of color PSNR. In addition, the visual appearance of the point clouds clearly increases with FSU. 
	
\end{abstract}


\section{Introduction and Related Work}
The increasing demand of capturing our environment for virtual and augmented reality applications~\cite{Mekuria_2017, Held_2012}, in automotive industry~\cite{Chen_2017, Geiger_2012}, in architecture, and archaeology~\cite{Mahmood_2020, Andriasyan_2020} drives the need for high-resolution point clouds. Point clouds are a versatile three-dimensional data type. In a point cloud, single points are captured using, e.g., a Light Detection and Ranging (LiDAR) sensor or an RGB-D camera such as the Microsoft Kinect~\cite{Han_2013}. For each point in a point cloud, the location in 3D space is stored. Moreover, each point may have an attribute assigned such as an intensity value or color information in RGB format. Such a set of many points forms a point cloud. As both, geometry and attribute, have to be stored for each point in a point cloud, this data type requires large storage capacities. However, many applications demand for high resolution point clouds. Therefore, point clouds often have to be upsampled artificially after acquisition.
\par The upsampling of point clouds applies to both, the geometry and the attributes of a point cloud. As a consequence of this, point cloud upsampling is generally separated into two steps, geometry upsampling and attribute upsampling. In the geometry upsampling part, we focus on retrieving the best location for the upsampled points whereas in the attribute upsampling part, we focus on precisely estimating the attribute at the upsampled positions. In literature, mainly the geometry upsampling part has been investigated so far.
\begin{figure}
	\includegraphics[width=\columnwidth]{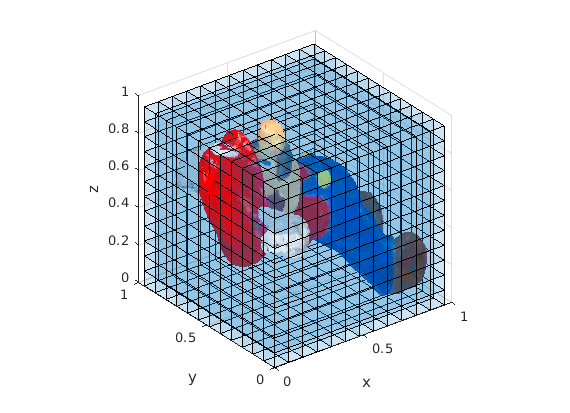}
	\caption{\label{fig:pcloud_partitioned}\texttt{Mario} point cloud is partitioned into blocks.}
	\vspace{-.3cm}
\end{figure}
\par Alexa et al. were the first to add additional points to a point cloud's surface \cite{Alexa_2001}. They initially investigated the problem of point cloud reconstruction. Point cloud reconstruction describes the process of estimating the surface of a point cloud in order to reconstruct missing areas. In \cite{Alexa_2001}, point set surfaces are presented for point cloud reconstruction. With the point set surfaces, an estimation of the point cloud's surface is established. In a subsequent step the estimated surface is sampled such that another representation of the surface is created. The sampling step size steers the accuracy and smoothness of the new surface representation. Thereby, Alexa et al. were the first to sample a point cloud's surface and thus, adding new points to the set of originally available points. Other approaches to point cloud reconstruction aim at solving an indicator function in three-dimensional space. The surface is then generated by isosurfacing the grid~\cite{Berger_2013}. Usually, these algorithms work on a regular grid or on octree. If the normal field agrees with the local derivation of the surface, the indicator function can be found by solving a Poisson equation~\cite{Kazhdan_2006}. Apart from point cloud reconstruction, point cloud upsampling was shown in Lipman et al. \cite{Lipman_2007}. They introduced a locally optimal projection (LOP). For this projection, a set of projected points is defined such that it minimizes the sum of weighted distances to the given point set. The LOP can also refine noisy data sets. Thus it is also applied for the removal of noise and outliers of raw scanned input data. The edge-aware resampling approach (EAR) by Huang et al. \cite{Huang_2013} incorporates normal vectors into the upsampling scheme. In this approach, the assumption is exploited that normal vectors of points in homogeneous areas that are far away from edges are more accurate than normal vectors in edge-like areas. Thus, the upsampling procedure starts within the homogeneous areas and continues with the upsampling progressively to the edge areas. Finally, the remaining regions are upsampled. EAR produces point sets with accompanying normal vectors. Normal vectors are also incorporated in the approach from Dinesh et al.~\cite{Dinesh_2019}. They assume locally smooth surfaces and thereby assume only small deviations between normal vectors of neighboring points. However, a major problem in point cloud processing is the missing knowledge regarding neighborhood relations. Dinesh et al. overcome this problem with a k-nearest-neighbor graph that connects the single points of a point cloud. The Euclidean distances are incorporated as a measure to determine the nearest neighbors. In addition, the graph holds weights that are determined based on the similarity of neighboring nodes, i.e., points of the point cloud. The upsampled points are inserted based on a Delaunay triangulation. Their locations are optimized during a refinement step. Therefore, the problem is reformulated as a minimization of a graph-total variation.
\par Since the development of PointNet in 2017 \cite{Charles_2017}, the processing of point cloud problems with neural networks gained much interest. The first network that performed point cloud upsampling was Point Cloud Upsampling Net (PU-Net)~\cite{Yu_2018_PUNet}. It is built upon PointNet++ \cite{Qi_2017}. PU-Net splits the input point cloud into smaller patches. These patches are used to train the multi-level features using hierarchically learning from PointNet++ \cite{Qi_2017}. The features from each level are subsequently interpolated and concatenated. As a result, embedded point features are generated. These are expanded and used for the three-dimensional coordinate reconstruction. For the training of the network, a joint loss function is incorporated that balances between a smooth surface and a uniform distribution of the points. Numerous further networks build upon PU-Net. Yifan et al. \cite{Yifan_2019} use PU-Net in a multi-step patch-based network (MPU-Net). The aim is to adapt the receptive field of the network. Yu et al. \cite{Yu_2018_ECNet} introduced the edge-aware consolidation network (EC-Net). It especially learns to extract edges as features during the training phase and mainly adds upsampled points in edge areas. Furthermore, also a generative adversarial network (GAN) approach was presented for point cloud upsampling by Li et al. \cite{Li_2019_PUGan} with PU-GAN. Zhang et al. \cite{Zhang_2019_DataDriven} do not follow a local patch-based approach but use the point cloud as a whole as input to their network. The clear disadvantage of this approach is that the point clouds always must have the same overall number of points in order to meet the requested input size of the network. PUGeoNet \cite{Qian_2020} does not learn the features in a three-dimensional domain. The three-dimensional surface is projected onto a two-dimensional plane first. The local parametrization for the transformation is learnt. Thereafter, the point cloud upsampling is pursued in the two-dimensional domain. Finally, the points are shifted back to the three-dimensional domain by a linear transformation. Meta-PU \cite{Ye_2021} is the first data-driven method that aims at upsampling a point cloud by an arbitrary scaling factor.
\par However, all these approaches hold drawbacks. The optimization-based approaches mainly rely on normal vectors. Unfortunately, normal vectors are not available for every point cloud. As the calculation of normal vectors is highly sensitive to noise and point clouds are often noisy due to their acquisition process, the calculated normal vectors are not accurate. Also the data-driven neural network based approaches hold drawbacks. They are mainly trained for distinct use cases such as a specific data set or scaling factor. Thus, the generalization to new unseen data sets might be a challenge. Therefore, Frequency-Selective Geometry Upsampling (FSGU) was introduced in~\cite{Heimann_2022_ICIP}. A model-based approach that estimates the object's surface block-based and iteratively with cosine basis functions. For the block partitioning, the points are assigned to solely one block and all points in one block are processed together. The model exploits the frequency selectivity principle which is further explained in the upcoming section. The underlying assumption is that the object's surface can be represented locally in terms of a finite number of basis functions. During model generation, the influence of the underlying frequency parts are estimated. The continuously estimated surface can then be sampled at new positions such that any arbitrary scaling factor can be achieved without the aid of normal vectors.
\par The presented methods for upsampling the point cloud geometry produce only the locations of the upsampled points. Thus, the missing attribute information has to be assigned to the upsampled points in a subsequent attribute upsampling step. Upsampling is a well-known problem for two-dimensional images and is often also referred to as single-image super-resolution. Numerous methods were developed for image upsampling \cite{Park_2003_SROverview, Dong_2016, Yang_2014}. A straight-forward approach is to use interpolation schemes such as bilinear or bicubic interpolation \cite{Park_2003_SROverview, Yang_2014}. These interpolation schemes are commonly implemented incorporating a triangulation scheme. For three-dimensional applications, this is a significant drawback as for some point locations an extrapolation is required. This occurs for example in cases of a concave object surface. Extrapolation is not possible for triangulation-based schemes as the interpolated point has to be surrounded by points located at the corners of a triangle. For extrapolation such a triangle cannot be built and thus the interpolation scheme fails to estimate an attribute for this point. Hence, the interpolation schemes from two-dimensional applications cannot be transferred directly to three-dimensional surfaces. Dinesh et al. \cite{Dinesh_2020} extended their graph-total variation approach also to color upsampling of point clouds. A first estimation for the RGB values is conducted as the mean of the surrounding color values. Thereafter, the estimation is refined. As for the geometry, they assume the neighborhood to be piecewise smooth, i.e., they assume a smooth color surface. With this assumption, they can once again reformulate the refinement as a minimization of a graph total variation term. Based on the frequency selectivity principle, Frequency-Selective Mesh-to-Mesh Resampling (FSMMR) was introduced in~\cite{Heimann_2021_MMSP}. The color attribute of a point cloud is represented in terms of a weighted superposition of basis functions and the frequency-selectivity principle is applied. For the model estimation, the three-dimensional surface of the object is projected into two-dimensional space. For the projection, a minimum spanning tree is established. It is based on the Euclidean distances between neighboring points. The projection into two-dimensional space is then conducted along the minimum spanning tree in order to cope for the object's extension in \(z\)-dimension. Next, the frequency model is established for the color attribute based on the projected coordinates. However, the minimum spanning tree has to be established for each block separately which is a complex process. Hence, we propose to simplify the projection by the incorporation of geometry information from the geometry upsampling in this work. 
\begin{figure}[t!]
	\scalebox{.75}{
%

\definecolor{colorflow}{rgb}{0.00000,0.44700,0.74100}%

\tikzset{%
	back group/.style={fill=yellow!20,rounded corners, draw=black!50, dashed, inner xsep=12pt, inner ysep=11pt, yshift=-5pt}
}

\tikzstyle{decision} = [diamond, draw, fill=white!15,
text width=4.5em, text badly centered, node distance=3cm, inner sep=0pt]
\tikzstyle{block} = [rectangle, draw, fill=white!15, 
text width = 20em, text centered, rounded corners, minimum height=2em] 
\tikzstyle{block2} = [rectangle, draw=colorflow, fill=white!15, line width = 2pt,
text width = 20em, text centered, rounded corners, minimum height=2em] 
\tikzstyle{blockk} = [rectangle, draw, fill=colorflow, 
text width = 20em, text centered, rounded corners, minimum height=2em] 
\tikzstyle{blockkk} = [rectangle, draw=white, fill=white!15, 
text width = 20em, text centered, rounded corners, minimum height=2em] 
\tikzstyle{blocksummarize} = [rectangle, draw, fill=white!15, line style=dashed,
text width = 20em, text centered, rounded corners, minimum height=2em] 
\tikzstyle{line} = [draw, -latex']
\tikzstyle{cloud} = [draw, ellipse,fill=white!20, node distance=5cm,
minimum height=2em]

\begin{tikzpicture}[node distance = 1.5cm, auto]
	\node [blockkk] (init) {Original signal};
	\node [block, below of=init] (residual) {Calculate residual $r^{(\nu)}(m,n)$};
	\node [block, below of=residual] (energy) {Calculate residual energy decrease $\Delta E^{(\nu)}$ for every basis function};
	\node [block, below of = energy] (selection) {Selection of best fitting basis function};
	\node [decision, below of=selection, node distance = 2.5cm] (stop) {Stopping criterium met?};
	\node [block, below of= stop, node distance = 2.5cm] (mesh) {Obtain signal at new points $ (m',n')$};
	\node [blockkk, below of=mesh, node distance = 1cm] (finish) {Final signal};
	
	\coordinate[right of=residual] (a1);  
	\coordinate[right of=stop] (e1); 
	
	\path [line] (init) -- (residual);
	\path [line] (residual)  -- (energy);
	\path [line] (energy) -- (selection);
	\path [line] (selection) -- (stop);
	\path [line] (stop) -| node [near start]{No}([xshift=3.0cm]e1) -- node[sloped, anchor=center, above, text width = 5cm](atest){Generated model $g^{(\nu)}(m',n')$}([xshift=3.0cm]a1) -- (residual);
	\path [line] (stop) -- node[near start]{Yes} (mesh);
	\path [line] (mesh) -- (finish);
	
\end{tikzpicture}
	}
	\caption{\label{fig:freq_sel} Frequency selectivity principle.}
	\vspace{-.3cm}
\end{figure}
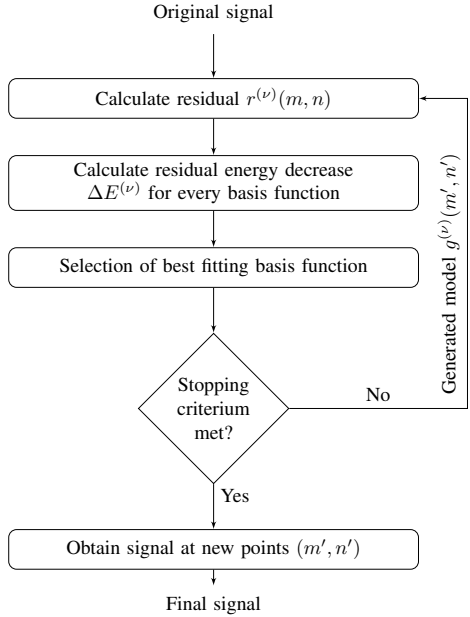
\vspace{-.1cm}
\par However, no point cloud upsampling scheme has been reported yet that solves both tasks, geometry and attribute upsampling in a single and joint scheme. Thus, we propose a joint geometry and attribute point cloud upsampling scheme in the following. Therefore, the point cloud is partitioned into blocks with overlapping support area for smoother results and less blocking artifacts. We establish a frequency model for surface extraction deployed for the geometry upsampling as in \cite{Heimann_2022_ICIP} and a frequency model for upsampling the corresponding attribute as in \cite{Heimann_2021_MMSP}. Therefore, the surface of the three-dimensional object is projected into a two-dimensional domain. For this transformation, the knowledge from the geometry upsampling about the surface is incorporated. Our approach uses the location information from the points and the attributes solely. No additional information such as normal vectors are required. Furthermore, the proposed algorithm can easily be adapted to new data sets and scaling factors.
\par The frequency selectivity principle is presented in the upcoming section. Thereafter, we present our proposed frequency-selective upsampling in Sec.~\ref{sec:method}. In Section \ref{sec:eval}, it follows the extensive evaluation of our proposed upsampling scheme. Finally, in Section~\ref{sec:con}, a conclusion is drawn.

\section{\label{sec:principle} Frequency Selectivity Principle}
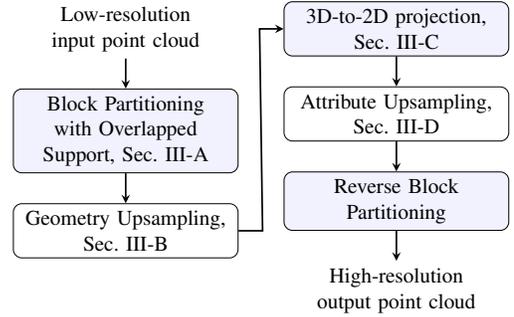
\begin{figure}[t!]
	\centering
	\scalebox{.8}{

\tikzstyle{block} = [rectangle, rounded corners, minimum width=3cm, minimum height=.5cm, text width=3.5cm, text centered, draw=black] 
\tikzstyle{block_col} = [rectangle, rounded corners, minimum width=3cm, minimum height=.5cm, text width=3.5cm, text centered, draw=black, fill=blue!5] 
\tikzstyle{hide} = [rectangle, minimum width=3cm, minimum height=.5cm, text width=3cm, text centered]
\tikzstyle{arrow} = [thick,->,>=stealth]

\begin{tikzpicture}
	\node[hide] (lowres) {Low-resolution input point cloud};
	\node[block_col, below=.5cm of lowres] (block) {Block Partitioning with Overlapped Support, Sec.~\ref{sec:blockpartitioning}};
	\node[block, below=.5cm of block] (geo) {Geometry Upsampling, Sec.~\ref{sec:geoup}};
	\node[block_col, right=of lowres] (proj) {3D-to-2D projection, Sec.~\ref{sec:3to2}};
	\node[block, below=.5cm of proj] (att) {Attribute Upsampling, Sec.~\ref{sec:att_up}};
	\node[block_col, below=.5cm of att] (rev) {Reverse Block Partitioning};
	\node[hide, below=.5cm of rev] (highres) {High-resolution output point cloud};
	
	\coordinate[right=.4cm of geo](a);
	\coordinate[left=.4cm of proj](b);
	
	\draw[arrow] (lowres) -- (block);
	\draw[arrow] (block) -- (geo);
	\draw[arrow] (geo.east) -- (a) -- (b) -- (proj.west);
	\draw[arrow] (proj) -- (att);
	\draw[arrow] (att) -- (rev);
	\draw[arrow] (rev) -- (highres);
	
\end{tikzpicture}
	}
	\caption{\label{fig:flow_joint} Joint Frequency-Selective Upsampling. Newly introduced steps are highlighted in blue.}
\end{figure}
The frequency selectivity principle has already been proven to be superior in several resampling~\cite{Heimann_2021_MMSP, Koloda_2017} , reconstruction~\cite{Heimann_2022_ICIP, Seiler_2015}  and extrapolation scenarios~\cite{Kaup_2005}. Therefore, the signal is first partitioned into blocks. The set of points in one block is denoted as \(\mathcal{A}\) and computed jointly. The model always follows the assumption that a signal \(f\) that is known at distinct positions \((m,n)\) with floating accuracy can be represented in terms of a weighted position of basis functions \(\varphi\), i.e., 
\begin{equation}
		\label{eq:beginning}
	f(m,n) = \sum_{k, l \in \mathcal{K}} c_{k, l} \varphi_{k, l}(m,n),
\end{equation}
where \(k,l\) denote the frequency indices of the basis functions from the set of available basis functions \(\mathcal{K}\) and \(c\) is the according expansion coefficient. With our model \(g\), we aim at estimating \eqref{eq:beginning} in an iterative process. Thus, we define the model to be 
\begin{equation}
	\label{eq:model}
	g^{(\nu)}(m,n) = g^{(\nu -1)}(m,n) + \hat{c}_{u,v} \varphi_{u, v}(m,n)
\end{equation}
with \((u,v)\) being the selected frequency indices in one iteration \(\nu\) and \(\hat{c}\) being the estimated expansion coefficient. In the beginning, the model is set to zero, i.e., \(g^{(0)}=0\). Any arbitrary type of basis functions can be incorporated into the model estimation process. We mainly incorporate cosine basis functions. These provide dense energy compaction such that a precise model can be found with a small number of iterations. Furthermore, the basis functions are real-valued which is advantageous for scattered input data. The difference between the original signal and the modeled signal is referred to as residual \(r\). Thus, 
\begin{equation}
	\label{eq:residual}
	r^{(\nu)}(m,n) = f(m,n) - g^{(\nu)} (m,n).
\end{equation} 
The task of the iterative model estimation procedure is to minimize the deviation between the original signal and the model, i.e., to minimize the residual as good as possible. For the minimization of the residual, the residual energy \(E\)
\begin{equation}
	\label{eq:resenergy}
	E^{(\nu)} = \sum_{(m,n)} w(m,n)\left(r^{(\nu)}(m,n)\right)^2
\end{equation}
is calculated in every iteration. Formulating the optimization problem as an energy holds the advantage that the minimization of the residual is independent of the sign of the residual. Furthermore, a spatial weighting function \(w(m,n)\)  is incorporated that steers the influence of every single point. The spatial weighting function is usually defined as a decaying isotropic window function. Thus, the center points have higher weights assigned than the points in the outer part of the block. Due to the higher weights in the center, the model estimates the centered points more accurately than the points in the outer areas as the smaller weights allow for a larger deviation from the original positions.
\par In the final step of each iteration, the best fitting basis function has to be selected. We select the basis function as the best fitting basis function in one iteration that reduces the residual energy the most. Thereby, closing the gap between original and model as good as possible. The best fitting basis function is defined by its two-dimensional frequency indices \((u,v)\) and thus, 
\begin{equation}
	\label{eq:bestfittingbf}
	{(u, v)} = \underset{{(k, l)}}{\mathrm{argmax}} \left( \Delta E_{k, l}^{(\nu)} w_{f}[k,l] \right).
\end{equation}
\begin{figure}[t]
	\centering
	\resizebox{.75\columnwidth}{!}{
		\input{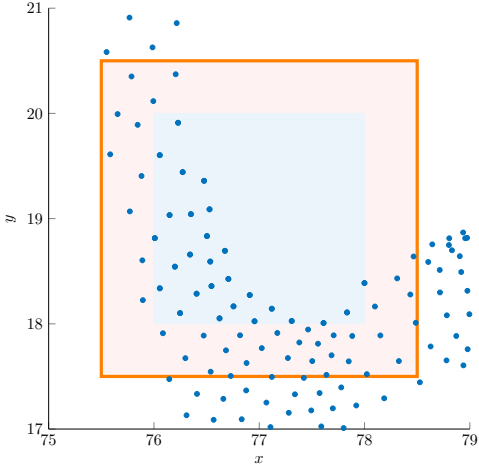}
	}
	\caption{\label{fig:blockpartitioning} Scattered input data that is partitioned into a core block (blue) of size \(N=2\) with support area (red) of \(M=0.5\). Together, they build a block (orange). Points on white background are not used for model estimation.}
	\vspace{-.3cm}
\end{figure}
During the maximization of the residual energy decrease, an additional spectral weighting function \(w_{f}\) is incorporated. The spectral weighting function remaps the assumption that the underlying signal is locally smooth. Hence, the signal mainly consists of low-frequency basis functions. High frequency functions tend to produce an oscillating signal that appears to be noisy. This behavior is usually not desired. Hence, the spectral weighting function is a smoothly decaying function that assigns higher weights to low-frequency basis functions and smaller weights to high-frequency functions. It is once again described as an isotropically decaying window function, i.e., 
\begin{equation}
	\label{eq:freqweighting}
	w_{f}[k,l] = \sigma^{\sqrt{k^2+l^2}}
\end{equation}
with decaying factor \(\sigma\). The isotropically decaying weighting function allows to include high frequencies if they are dominant in the signal while preserving a mainly low frequency signal. We now choose the best fitting basis function in iteration \(\nu\) and add this function to our estimated model from the previous iteration step \((\nu-1)\). However, the expansion coefficient of the chosen basis function has to be determined. Therefore, we follow the approach from \cite{Kaup_2005} and set the derivative of the residual energy \(E^{(\nu)}\) to zero. Thus, the expansion coefficient is given as
\begin{equation}
	\label{eq:expansioncoefficient}
	\hat{c}^{(\nu)}_{(k, l)} = \frac{\sum_{(m,n)\in \mathcal{A}} r^{(\nu-1)}(m,n) \varphi_{(k,l)}(m,n)}{\sum_{(m,n)\in \mathcal{A}} w(m,n)(\varphi_{(k, l)}(m,n))^2}.
\end{equation}
If the model estimation is finished, the signal \(f'\) can be retrieved at new positions \((m',n')\), i.e., 
\begin{equation}
	f'(m',n') = \sum_{k, l \in \mathcal{K}} \hat{c}_{k, l} \varphi_{k, l}(m',n').
\end{equation}
The frequency-selective model estimation process is summarized in Fig.~\ref{fig:freq_sel}.

\section{Proposed Frequency-Selective Upsampling}
\label{sec:method}
In this work, we propose Frequency-Selective Upsampling (FSU), a point cloud upsampling scheme that can process both, geometry and attribute upsampling. Geometry and attribute upsampling are conducted jointly in a sequential manner. An overview of FSU is given in Fig.~\ref{fig:flow_joint}. Newly introduced steps are highlighted in blue.
Our proposed joint sequential point cloud upsampling scheme first partitions the point cloud into local blocks with overlapping support area. Moreover, it uses geometry information from the geometry upsampling of the point cloud in the upsampling of the attribute in the conducted projection step. 

\subsection{Block Partitioning \label{sec:blockpartitioning}}
The block partitioning is a crucial part for the proposed frequency models as the quality of the estimated model highly depends on the underlying original points. In particular, the block partitioning is only conducted once and used for geometry and attribute upsampling. We propose to partition the three-dimensional volume into blocks with an equal length of side of \(N\times N\times N\). These blocks form the core block. An exemplary three-dimensional block partitioning of the \texttt{Mario} point cloud is shown in Fig.~\ref{fig:pcloud_partitioned}. The point cloud is normalized in all three dimensions. The shown blocks hold a length of side of \(N=\frac{8}{100}\). However, research for two-dimensional applications has shown that an additionally added support area around the core block may increase the final quality~\cite{Koloda_2017, Kaup_2005}. Thus, we propose to add an additional support area for overlapped support. Therefore, the incorporated block size for model estimation is expanded to \( (N+2M) \times (N+2M) \times (N+2M) \) with \(M\) as the margin of the support area. Nevertheless, additional points in the upsampling procedure are only inserted in the area of the core block of \(N\times N\times N\). An example of the block partitioning is shown in Fig.~\ref{fig:blockpartitioning}. Here an excerpt of \texttt{Mario} is shown in two dimensions for better visibility. The scattered points are located on arbitrary coordinates with floating accuracy. We conduct a block partitioning for \(N=\frac{2}{100}\). For better demonstration, the axes are expanded by a factor \(100\) such that \(N=2\) here. The core block covers the interval \(x=[76, 78]\) and \(y=[18, 20]\). This area is highlighted in blue. All points in this area are assigned to the core block. In addition, a support area with a margin of \(M=0.5\) is added. The support area is shown in red. Hence, all points that are located in the interval \(x=[75.5, 78.5]\) and \(y=[17.5, 20.5]\) are considered for model estimation. This bears the advantage of a smoother model especially in the border regions and smoother transitions from one block to the other. The partitioning process is conducted for all blocks. Thus, one point might be assigned to the support area of none, one or more blocks and at the same time it must be a member of exactly one core block. During this assignment, each point remains on its original location. Thus, this block partitioning procedure may not be confused with a voxelization that is often pursued as a preprocessing for point cloud processing algorithms. If the block partitioning is finished, one block after the other is handed over to the geometry upsampling step that estimates the underlying smooth object's surface in this one block. In the following, we refer to the joint set of core block and support area as a block. The points within one block form the set of points $\mathcal{A}$.
\begin{figure*}[t!]
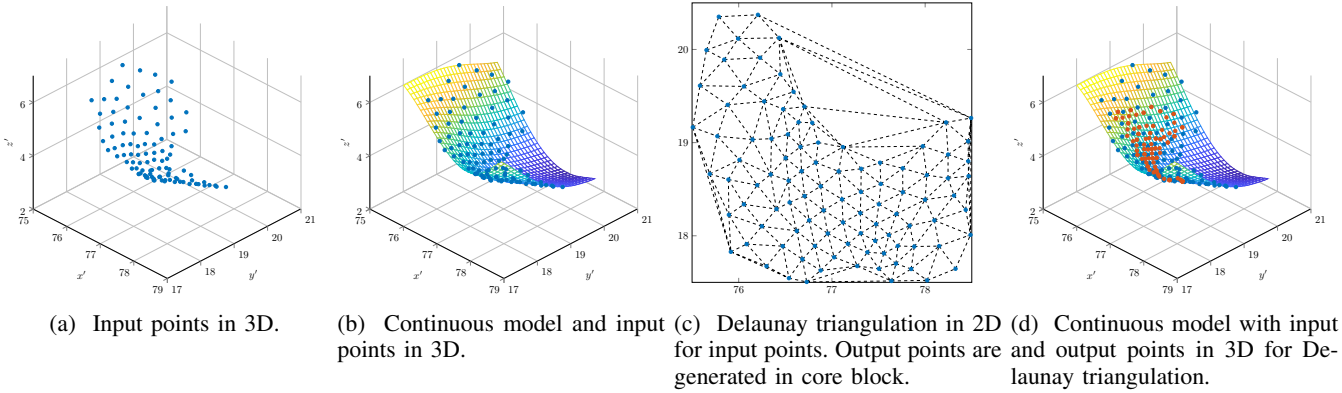

	\begin{subfigure}[t]{.24\textwidth}
		\scalebox{.4}{
%
\definecolor{mycolor1}{rgb}{0.00000,0.44700,0.74100}%
\begin{tikzpicture}

\begin{axis}[%
width=3.504in,
height=3.566in,
at={(0.588in,0.481in)},
scale only axis,
xmin=75,
xmax=79,
tick align=outside,
ymin=17,
ymax=21,
zmin=2,
zmax=7,
view={44.9}{36.4},
axis background/.style={fill=white},
title style={font=\bfseries},
axis x line*=bottom,
axis y line*=left,
axis z line*=left,
xmajorgrids,
ymajorgrids,
zmajorgrids,
xlabel=\(x'\),
ylabel=\(y'\),
zlabel=\(z'\),
legend style={at={(1.03,1)}, anchor=north west, legend cell align=left, align=left, draw=white!15!black}
]

\addplot3[only marks, mark=*, mark options={}, mark size=1.5811pt, color=mycolor1, fill=mycolor1] table[row sep=crcr]{%
x	y	z\\
75.5074996948242	19.1642990112305	6.3161997795105\\
75.5845031738281	19.6110000610352	5.87709999084473\\
75.6546936035156	19.9934997558594	5.51219987869263\\
75.6897964477539	18.6637992858887	6.16829967498779\\
75.7711029052734	19.0685997009277	5.73819971084595\\
75.7869033813477	20.3511009216309	5.10270023345947\\
75.8452987670898	19.891300201416	5.08470010757446\\
75.8818969726562	19.4053001403809	5.29010009765625\\
75.8899002075195	18.6036014556885	5.66769981384277\\
75.8945007324219	18.2249984741211	5.90360021591187\\
75.9154968261719	17.8279991149902	6.13609981536865\\
75.9940032958984	20.1166000366211	4.69250011444092\\
76.0081024169922	18.8144989013672	5.28090000152588\\
76.0551986694336	19.603099822998	4.78350019454956\\
76.0555038452148	18.3376007080078	5.45860004425049\\
76.0851058959961	17.9105987548828	5.68660020828247\\
76.1483993530273	19.0332984924316	4.8371000289917\\
76.1987991333008	18.5433006286621	5.05109977722168\\
76.2063980102539	20.3721008300781	4.21129989624023\\
76.2292022705078	19.9109992980957	4.32929992675781\\
76.2481002807617	18.1017990112305	5.29610013961792\\
76.272705078125	19.4419994354248	4.41449975967407\\
76.2984008789062	17.6735000610352	5.46020030975342\\
76.3427963256836	18.6585998535156	4.62470006942749\\
76.3510971069336	19.0424003601074	4.39820003509521\\
76.404899597168	18.286600112915	4.91480016708374\\
76.4334030151367	20.1245002746582	3.81180000305176\\
76.4468002319336	19.7341003417969	3.97420001029968\\
76.4724044799805	17.8882999420166	5.09619998931885\\
76.4760971069336	19.3595008850098	4.04930019378662\\
76.5034942626953	18.8341007232666	4.21749973297119\\
76.5278015136719	19.0882987976074	4.05369997024536\\
76.5348052978516	18.5918006896973	4.31839990615845\\
76.5386962890625	17.5450992584229	5.19409990310669\\
76.5461959838867	18.3586006164551	4.55179977416992\\
76.5883026123047	19.5454998016357	3.72960019111633\\
76.622802734375	18.0520992279053	4.76490020751953\\
76.6419982910156	19.2192993164062	3.79620003700256\\
76.6627044677734	18.9547004699707	3.91420006752014\\
76.6740951538086	18.6940002441406	4.05949974060059\\
76.6838989257812	17.7481002807617	4.8914999961853\\
76.7052001953125	19.386999130249	3.5939998626709\\
76.7067031860352	18.426700592041	4.21059989929199\\
76.7297058105469	17.5044994354248	5.00649976730347\\
76.7554016113281	18.1660003662109	4.43149995803833\\
76.7886962890625	19.2126998901367	3.60610008239746\\
76.8173980712891	17.8908996582031	4.62660026550293\\
76.8273010253906	18.7856998443604	3.87170028686523\\
76.8525009155273	19.0004997253418	3.70789980888367\\
76.8791046142578	18.5314998626709	3.98230004310608\\
76.8792953491211	17.6275997161865	4.79199981689453\\
76.9096984863281	18.2733993530273	4.13590002059937\\
76.9552001953125	18.0244998931885	4.35659980773926\\
77.0240020751953	17.769100189209	4.57849979400635\\
77.0718002319336	18.6912994384766	3.77199983596802\\
77.1085052490234	18.3914012908936	3.90359973907471\\
77.1176986694336	18.9519996643066	3.59640002250671\\
77.118896484375	18.1436996459961	4.08080005645752\\
77.171501159668	17.913200378418	4.34219980239868\\
77.2719039916992	17.6747989654541	4.5556001663208\\
77.3046951293945	18.2432994842529	3.88839983940125\\
77.3085021972656	18.027099609375	4.09569978713989\\
77.310905456543	18.4711990356445	3.75629997253418\\
77.3137969970703	18.718900680542	3.64909982681274\\
77.3805999755859	17.8227996826172	4.35060024261475\\
77.4637985229492	17.9458999633789	4.15530014038086\\
77.4731979370117	18.1254005432129	3.95959997177124\\
77.5032958984375	17.6459999084473	4.519700050354\\
77.5037002563477	18.3376007080078	3.82999992370605\\
77.5165939331055	18.7842998504639	3.59710001945496\\
77.5240020751953	18.5564002990723	3.71869993209839\\
77.5578994750977	17.8110008239746	4.32719993591309\\
77.6106033325195	18.0074996948242	4.11279964447021\\
77.638298034668	17.5149002075195	4.69670009613037\\
77.6867980957031	17.701000213623	4.51040029525757\\
77.6918029785156	18.2341003417969	3.98579978942871\\
77.7065963745117	17.8908996582031	4.31220006942749\\
77.7588043212891	18.724100112915	3.68250012397766\\
77.7602005004883	18.4673004150391	3.853600025177\\
77.8338012695312	18.1084003448486	4.20279979705811\\
77.8498001098633	17.6432991027832	4.74900007247925\\
77.8832015991211	17.8844013214111	4.5\\
77.9738998413086	18.9612007141113	3.5884997844696\\
77.9991989135742	18.3873996734619	4.07070016860962\\
78.0177001953125	18.6506996154785	3.82790017127991\\
78.0219955444336	17.5214996337891	5.0890998840332\\
78.0986022949219	18.1646995544434	4.41949987411499\\
78.1501998901367	17.8895988464355	4.80590009689331\\
78.2244033813477	19.2166996002197	3.52770018577576\\
78.2452011108398	18.6180000305176	3.99940013885498\\
78.2455978393555	18.8721008300781	3.77149987220764\\
78.3097991943359	18.4320011138916	4.29460000991821\\
78.3258972167969	17.6459999084473	5.28229999542236\\
78.434700012207	18.2786998748779	4.67560005187988\\
78.4610977172852	19.013599395752	3.77870011329651\\
78.4618988037109	18.8001003265381	3.97650003433228\\
78.4655990600586	18.6401996612549	4.18910026550293\\
78.4864044189453	18.0101013183594	5.05089998245239\\
78.492805480957	19.2637996673584	3.62109994888306\\
};

\end{axis}
\end{tikzpicture}%
}
		\caption{\label{Fig:mario_input} Input points in 3D.}
	\end{subfigure}
	\begin{subfigure}[t]{.24\textwidth}
		\scalebox{.4}{\input{figures/Mario_model_and_input_2}}
		\caption{\label{Fig:mario_modelinput} Continuous model and input points in 3D.}
	\end{subfigure}
	\begin{subfigure}[t]{.24\textwidth}
		\scalebox{.4}{\input{figures/Mario_delaunay}}
		\caption{\label{Fig:mario_inputoutput} Delaunay triangulation in 2D for input points. Output points are generated in core block.}
	\end{subfigure}
	\begin{subfigure}[t]{.24\textwidth}
		\scalebox{.4}{\input{figures/Mario_model_and_input_and_output_2}}
		\caption{\label{Fig:mario_modelinputoutput} Continuous model with input and output points in~3D for Delaunay triangulation.}
	\end{subfigure}
	\caption{\label{Fig:modelling}Model generation process for a block of the \texttt{Mario} point cloud. Blue points are the original points, red points are the upsampled points. The continuous model is depicted as a mesh plot.}
	\vspace{-.2cm}
\end{figure*}   
\vspace{-.2cm}
\subsection{Geometry Upsampling}
\label{sec:geoup}
In the geometry upsampling of a point cloud, points have to be added to the original point cloud. Therefore, the points' locations have to be determined first. These have to satisfy two essential requirements. First, the points should fit well in the surface of the object. And second, the points' distribution should approximately follow a uniform distribution such that the newly added points are not located directly next to original points. We follow the approach as described in~\cite{Heimann_2022_ICIP}. We assume a point cloud's surface to be locally smooth and representable in terms of a function within one block. One exemplary block is depicted in Fig.~\ref{Fig:mario_input}. At first sight, the block just seems to contain a set of loose points. However, we will estimate a smooth underlying surface in the following. Subsequently, we place the additional points on this surface. \par 
In a first step, it has to be decided in which dimensions the surface model is estimated. Therefore, the dimension that yields the smallest variance is selected. This bears two advantages. First, the probability that the surface is a closed form in the dimension with smallest extension is low. Second, if the first assumption is false, the introduced error is rather small as only one surface is estimated as averaging surface in between the other surfaces that build a closed form. Thus, the modeled dimension \(z'\) is
\begin{equation}
	\label{eq:assign_vars}
	z' = \min\left\lbrace \text{Var}\{x\}, \text{Var}\{y\}, \text{Var}\{z\}\right\rbrace
\end{equation}
with \(x, y\) and \(z\) being the three-dimensional coordinates of the point cloud's points.
The main assumption that we follow for geometry upsampling is that the underlying surface in one block can be represented in terms of a function. We assume the function to be a weighted superposition of basis functions \(\varphi\), i.e., 
\begin{equation}
	z'=f(x',y') = \sum_{k, l \in \mathcal{K}} c_{k, l} \varphi_{k, l}[x',y'].
\end{equation}
This equation is closely related to \eqref{eq:beginning}, if we select \(f(m,n)=f(x',y')\) with \(m=x'\) and \(n=y'\). Thus, the model estimation is conducted along the explained steps from Sec.~\ref{sec:principle}. The resulting continuous model of the block is depicted in Fig.~\ref{Fig:mario_modelinput}.

Finally, the upsampled coordinates \(\hat{z}'\) are determined based on the upsampled points \((\hat{x}',\hat{y}')\) according to
\begin{equation}
		\hat{z}'=f(\hat{x}',\hat{y}') = \sum_{k, l \in \mathcal{K}} 	\hat{c}_{k, l} \varphi_{k, l}(\hat{x}',\hat{y}').
\end{equation}
The upsampled points \((\hat{x}',\hat{y}')\) are gained by a Delaunay triangulation of the points \((x',y')\). This computation can be conducted in parallel to the model generation procedure. The new points are inserted in the middle of the triangle edges as shown in Fig.~\ref{Fig:mario_inputoutput}. Thereby, high upsampling factors can be achieved with one triangulation. The final result of the block is given in Fig.~\ref{Fig:mario_modelinputoutput}. It is clearly visible, that the blue input points are located from one block border to the other, whereas the orange output points are only located in the core block. This is due to the incorporated support area. In this case a margin of \(0.5\) is taken as support for a core block of size \(2\times 2\times 2\).

\subsection{Proposed Projection}
\label{sec:3to2}
The attribute upsampling step describes the process of estimating the attribute's value at the upsampled positions determined as in Sec.~\ref{sec:geoup}. An attribute can be anything in point clouds such as color, intensity or normal vectors. 
\par As a starting point, we use the sparse frequency model approach from~\cite{Heimann_2021_MMSP}. We assume the point cloud object's surface to be a two-dimensional plane in three-dimensional space and thus, follow the approach to project the surface into a two-dimensional space. In~\cite{Heimann_2021_MMSP}, the Euclidean distances between the points are measured and incorporated as weights of a graph that is minimized to a minimum spanning tree. Following this tree, the points are mapped to a two-dimensional space. The transformation is conducted for each block. As the calculation of the euclidean distances and the generation of the minimum spanning tree is a time-consuming and complex process, we propose to exploit knowledge from our geometry upsampling which was presented in the section before. 
Therefore, we propose to simplify the 3D to 2D conversion. For geometry upsampling, the geometry is rotated such that the dimension with smallest variance is modeled. This bears the advantage that the standard deviation in \(z'-\)direction is small. Exploiting this, we can directly project the three dimensional points into two dimensions along the axis of the smallest dimension. Thus, 

\begin{center}
	\begin{tabular}{p{.3\columnwidth}c p{.3\columnwidth}}
		\vspace{-1cm}
		\begin{equation}
				\label{eq:x3_to_x2}
			x'_{2D} = x'
		\end{equation}
	\vspace{-1cm}
		&
		and 
		&
		\vspace{-1cm}
		\begin{equation}
				\label{eq:y3_to_y2}
			y'_{2D} = y'.
		\end{equation}
	\vspace{-1cm}
	\end{tabular}
\end{center}

Next, the two-dimensional points are used for the estimation of the attribute's frequency model. 
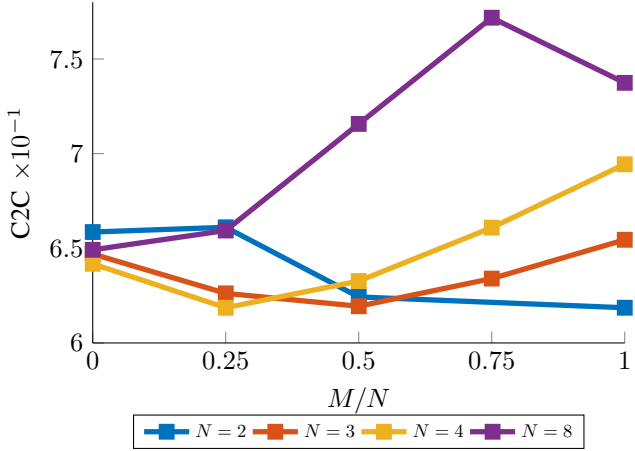
\begin{figure}
%
\definecolor{mycolor1}{rgb}{0.00000,0.44700,0.74100}%
\definecolor{mycolor2}{rgb}{0.85000,0.32500,0.09800}%
\definecolor{mycolor3}{rgb}{0.92900,0.69400,0.12500}%
\definecolor{mycolor4}{rgb}{0.49400,0.18400,0.55600}%
\begin{tikzpicture}

\begin{axis}[%
	width=.39\textwidth,
height=.25\textwidth,
at={(1.011in,0.642in)},
scale only axis,
xmin=0,
xmax=1,
xtick distance=0.25,
ymin=6,
ymax=7.8,
ytick distance=0.5,
xlabel=\(M/N\),
axis background/.style={fill=white},
axis x line*=bottom,
axis y line*=left,
ylabel = {C2C \(\times 10^{-1}\)},
legend columns=4,
legend style={at={(0.5,-0.21)},anchor=north, nodes={scale=0.7, transform shape}}
]
\addplot [color=mycolor1, line width=2pt, mark=cube*]
  table[row sep=crcr]{%
0	6.586\\
0.25	6.611\\
0.5	6.243\\
1	6.186\\
};
\addlegendentry{\(N=2\)}

\addplot [color=mycolor2, line width=2pt, mark=cube*]
  table[row sep=crcr]{%
0	6.47\\
0.25	6.262\\
0.5	6.194\\
0.75	6.34\\
1	6.545\\
};
\addlegendentry{\(N=3\)}

\addplot [color=mycolor3, line width=2pt, mark=cube*]
  table[row sep=crcr]{%
0	6.418\\
0.25	6.186\\
0.5	6.327\\
0.75	6.609\\
1	6.944\\
};
\addlegendentry{\(N=4\)}

\addplot [color=mycolor4, line width=2pt, mark=cube*]
  table[row sep=crcr]{%
0	6.492\\
0.25	6.594\\
0.5	7.157\\
0.75	7.718\\
1	7.374\\
};
\addlegendentry{\(N=8\)}

\end{axis}

\end{tikzpicture}%
	\caption{\label{fig:geo_blkSize}Geometry results in terms of C2C similarity for different block sizes $(N)$ and support margins ($M$). Support margin is given relative to block size ($M/N$).}
	\vspace{-.2cm}
\end{figure}
\subsection{Attribute Upsampling}
\label{sec:att_up}

The attribute information is assumed to be modeled with frequencies. As explained in \cite{Heimann_2021_MMSP}, the color signal \(f_a\) can be modeled as a weighted superposition of basis functions \( \varphi\) from the set of available basis functions \(\mathcal{K}\) according to 
\begin{equation}
	\label{eq:att_beginning}
	f_a(x'_{2D}, y'_{2D}) = \sum_{k,l \in \mathcal{K}} c_{k,l} \varphi_{k,l}(x'_{2D}, y'_{2D}).
\end{equation}
Once again, this assumption is closely related to Eq.~\eqref{eq:beginning}. Hence, for attribute upsampling we can formulate \(f(m,n) = f_a(x'_{2D}, y'_{2D}) \) with \(m=x'_{2D}\) and \(n=y'_{2D}\). During the model estimation process, a continuous model of the attribute of the point cloud for one block is estimated. Hence, the resulting model gives a continuous estimation of the point cloud's attribute. To assign a proper attribute information to the upsampled points \((\hat{x}'_{2D},\hat{y}'_{2D})\), the estimated model is evaluated for 

\begin{equation}
	f_a(\hat{x}'_{2D},\hat{y}'_{2D}) = \sum_{k,l \in \mathcal{K}} 	\hat{c}_{k,l} \varphi_{k,l}(\hat{x}'_{2D},\hat{y}'_{2D}).
\end{equation}
\par Finally, the additionally computed points with its geometric location and the associated color attribute are remapped in the three-dimensional space of the original point cloud. Thus, the equations \eqref{eq:assign_vars}, \eqref{eq:x3_to_x2} and \eqref{eq:y3_to_y2} have to be reversed. The 3D high-resolution point cloud in geometry and color is the final result. 

\vspace{-.1cm}
\section{Evaluation}
\label{sec:eval}

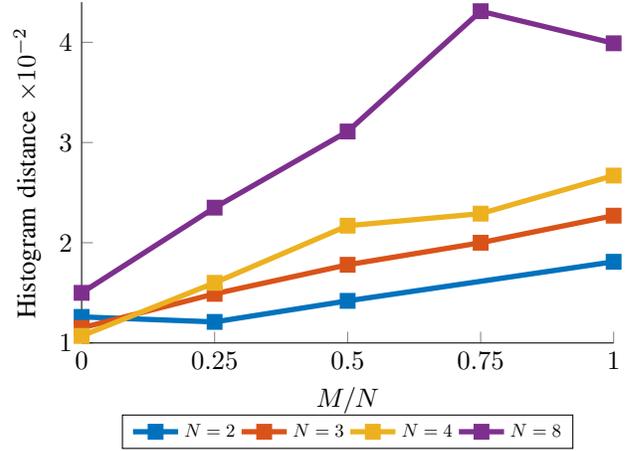
\begin{figure}
%
\definecolor{mycolor1}{rgb}{0.00000,0.44700,0.74100}%
\definecolor{mycolor2}{rgb}{0.85000,0.32500,0.09800}%
\definecolor{mycolor3}{rgb}{0.92900,0.69400,0.12500}%
\definecolor{mycolor4}{rgb}{0.49400,0.18400,0.55600}%
\begin{tikzpicture}

\begin{axis}[%
	width=.39\textwidth,
height=.25\textwidth,
at={(1.011in,0.642in)},
scale only axis,
xmin=0,
xmax=1,
xtick distance=0.25,
ymin=1,
ymax=4.4,
xlabel=\(M/N\),
axis background/.style={fill=white},
axis x line*=bottom,
axis y line*=left,
ylabel = {Histogram distance  \(\times 10^{-2}\)},
legend columns=4,
legend style={at={(0.5,-0.21)},anchor=north, nodes={scale=0.7, transform shape}}
]
\addplot [color=mycolor1, line width=2pt, mark=cube*]
  table[row sep=crcr]{%
0	1.26\\
0.25	1.21\\
0.5	1.42\\
1	1.81\\
};
\addlegendentry{\(N=2\)}

\addplot [color=mycolor2, line width=2pt, mark=cube*]
  table[row sep=crcr]{%
0	1.15\\
0.25	1.49\\
0.5	1.78\\
0.75	2\\
1	2.27\\
};
\addlegendentry{\(N=3\)}

\addplot [color=mycolor3, line width=2pt, mark=cube*]
  table[row sep=crcr]{%
0	1.07\\
0.25	1.6\\
0.5	2.17\\
0.75	2.29\\
1	2.67\\
};
\addlegendentry{\(N=4\)}

\addplot [color=mycolor4, line width=2pt, mark=cube*]
  table[row sep=crcr]{%
0	1.5\\
0.25	2.35\\
0.5	3.11\\
0.75	4.31\\
1	3.99\\
};
\addlegendentry{\(N=8\)}

\end{axis}

\end{tikzpicture}%
	\caption{\label{fig:col_blkSize}Color results in terms of histogram distance \cite{Viola_2020} for different block sizes ($N$) and support margins ($M$). Support margin is given relative to block size ($M/N$).}
	\vspace{-.3cm}
\end{figure}

For the evaluation of our proposed joint geometry and attribute upsampling scheme, we show evaluations for both, geometry and color, separately. In addition, visual examples of the joint upsampling scheme are shown. For the evaluations the \textit{3DColorMesh} dataset \cite{3DColMesh} is incorporated. This dataset contains colored point clouds with 40,000 to 200,000 points. 
\begin{table*}
	\caption{\label{tab:res_geo} Geometry results for all point clouds from the \textit{3DColorMesh} dataset for scaling factor of 4 in terms of P2P and P2C errors~\cite{Tian_2017} and C2C similarity~\cite{Alexiou_2020}. Best qualities are given in bold. Arrows indicate higher values are better \(\uparrow\) and smaller values are better \(\downarrow\), respectively. }
	\resizebox{\textwidth}{!}{%
		\begin{tabular}{|p{.18\textwidth}|C{.05\textwidth}|C{.05\textwidth}|C{.05\textwidth}|C{.05\textwidth}||C{.05\textwidth}|C{.05\textwidth}|C{.05\textwidth}|C{.05\textwidth}||C{.05\textwidth}|C{.05\textwidth}|C{.05\textwidth}|C{.05\textwidth}|}
			\hline
			& \multicolumn{4}{c||}{P2P $\times 10^{-3}$  \(\downarrow\)} & \multicolumn{4}{c||}{P2C $\times 10^{-3}$  \(\downarrow\)} & \multicolumn{4}{c|}{C2C$\times 10^{-1}$  \(\uparrow\)} \\
			\hline
			Point Cloud			& PU & EC &  FSGU & FSU&   PU & EC &  FSGU & FSU &  PU & EC &  FSGU & FSU \\
			\hline
			\hline
			\texttt{4armsMonstre} 	 & 8.9 & 4.5 & \textbf{3.1} & 3.3  & 7.2 & 2.9 & \textbf{2.0 }& \textbf{1.9}  &4.6 & 3.7 & 5.4 & \textbf{5.6} \\
			\hline
			\texttt{Asterix} 			 & 10.1  & 4.6 & \textbf{3.4} &  3.5 &  8.1 & 2.8 &\textbf{1.9} & \textbf{1.9}  & 4.5 & 3.6  & 4.6 & \textbf{4.9} \\
			\hline
			\texttt{CableCar}			 & 11.1  &  2.4 & \textbf{1.7} & 1.9  & 10.4 & 1.3 & 0.8 & \textbf{0.6}  & 5.0 & 3.7 & 7.3& \textbf{7.6} \\ 
			\hline
			\texttt{Dragon}				 & 11.7 & 2.1 & \textbf{1.6} & 1.9  & 11.2 & 6.9 & 0.6 & \textbf{0.5}  & 4.7 & 3.8 & 7.8 & \textbf{8.2} \\
			\hline
			\texttt{Duck}					& 11.4 & 5.3 & \textbf{2.9} & 3.3  & 9.3 & 2.1 & \textbf{0.7} & \textbf{0.7}  & 5.5 & 3.7 & 8.2 & \textbf{8.5} \\
			\hline
			\texttt{GreenDinosaur}	& 8.5 & 3.8 & \textbf{2.8} & 2.9  & 6.8 & 2.0 & \textbf{1.5} & \textbf{1.5}  & 4.1 & 3.7 & 4.8 & \textbf{4.9}\\
			\hline
			\texttt{GreenMonstre}	 & 9.3 & 2.3 & \textbf{1.8} & 1.9   & 8.4 & \textbf{0.9} & \textbf{0.9} & \textbf{0.9}  & 4.0 & 3.6 & \textbf{4.7} & \textbf{4.7} \\
			\hline
			\texttt{Horse}				& 7.9 & 3.0 & \textbf{2.7} & 2.8  &  6.4 & 1.7 & \textbf{1.7} &  \textbf{1.7}  &4.3 &3.7 & \textbf{5.4} & 5.2 \\
			\hline
			\texttt{Jaguar}				 & 11.4 & 1.7 & \textbf{1.3} & 1.5   & 11.1 & 0.6 & \textbf{0.5} &  \textbf{0.5} & 5.2 &3.7 & 7.7 & \textbf{8.0} \\
			\hline
			\texttt{LongDinosaur} 		& 12.5 & 1.5 & \textbf{1.2} & 1.4  & 12.2 & \textbf{0.5} & \textbf{0.5} &  \textbf{0.5} &5.1 &4.2 & 7.8 & \textbf{8.0} \\
			\hline
			\texttt{Man}					& 11.2 & 6.2 & \textbf{3.1} & 3.3  &  8.6 & 2.6 & \textbf{1.1} &  \textbf{1.1}  & 4.8 &3.8 & 6.6 & \textbf{6.9} \\
			\hline
			\texttt{Mario}					& 11.6 & 1.7 & \textbf{1.3} & 1.5  & 11.1 & 0.6 & \textbf{0.6} & \textbf{0.6}  & 5.0 & 3.7 & 7.7 & \textbf{7.8}\\
			\hline
			\texttt{MarioCar}			 & 11.0 & 1.9 & \textbf{1.4} & 1.7  & 10.6 & 0.7 & 0.6 & \textbf{0.5}   & 4.9 & 3.7& 7.7 & \textbf{8.0}\\
			\hline
			\texttt{PokemonBall}	 & 9.7 & 8.0 &4.6 & \textbf{4.5}  & 6.5 & 3.5 & 2.7 & \textbf{2.1}  & 4.4 & 3.4 & 4.5& \textbf{4.9}\\
			\hline
			\texttt{Rabbit}				 & 8.7 & 3.4& \textbf{2.7} & 2.9   & 7.3 &  2.1 & \textbf{1.8} &  \textbf{1.8} &  4.5 & 3.6 & \textbf{5.8 }& 5.6 \\
			\hline
			\texttt{RedHorse}		& 10.8 & 1.8 & \textbf{1.5} & 1.8   &10.3 & 0.7 & \textbf{0.7} & \textbf{0.7}  & 4.7 & 3.7 & 7.4 & \textbf{7.7}\\
			\hline
			\texttt{Statue} 			& 8.5 & 3.6 & \textbf{2.8} & 3.0  & 7.0 & 2.3 & 1.9 & \textbf{1.8}   & 4.6 & 3.8& 5.7 & \textbf{5.8} \\
			\hline
			\hline
			Average 			 & 10.2 & 3.5 & \textbf{2.3} & 2.5  & 8.9 & 1.9 & 1.2 & \textbf{1.1}  & 4.7 & 3.7 & 6.4 & \textbf{6.6} \\
			\hline
		\end{tabular}
	}
		\vspace{-.3cm}
\end{table*}

\vspace{-.2cm}
\subsection{Metrics}
The quality of the upsampled point clouds are determined for both, geometry and color, separately. For each upsampling part, different metrics are applied. 
\subsubsection{Geometry}
The evaluation of the geometric shape of the upsampled point cloud is usually done with respect to the original point cloud. Therefore, the original point cloud serves as a reference. For each point in the upsampled point cloud, the nearest neighbor in the reference is searched. The deviations are summed up and normalized for all points such that the overall point-to-point (P2P) error is determined. Hence, the P2P error is the normalized sum of the error vectors being the smallest distance between the points in the reference and the upsampled point cloud~\cite{Tian_2017}.
\par A related method is to determine the point-to-plane (P2C) error. Therefore, the same procedure as for the point-to-point error is followed with the difference that not the direct deviation between the upsampled point and the nearest point in the reference is taken but the difference along the normal of the upsampled point is taken. Once again, these differences are summed up and normalized such that the overall point-to-plane error is determined~\cite{Tian_2017}.
\par As a third metric, the plane-to-plane (C2C) angular similarity is determined. In this metric, a plane is estimated in both, the reference and the upsampled point cloud. Then, the angular similarity between the two planes is determined. Thereby, the visual degradations of a processed point cloud can be predicted more accurately. The plane-to-plane similarity metric is determined according to the implementation of Alexiou et al.~\cite{Alexiou_2020}.
\subsubsection{Attribute}
\label{sec:metric_att}
A great challenge in the joint upsampling of the geometry and attribute of a point cloud is the determination of the final attribute quality as it is highly affected by geometric distortions. Thus, we decided to evaluate geometry and attribute separately. In order to have ground truth data available, we first downsample the original point cloud randomly. The downsampled points incorporate both, geometry and color information and thus, serve as the original points. From the remaining points, only the geometry information is kept such that a possible geometrical distortion is not affecting the attribute quality during the evaluation. The attribute information of these points is determined following the proposed algorithm described in Sec.~\ref{sec:method}. Thus, the overall point cloud is separated into blocks. Next, each block is rotated according to the geometrical variances. The geometrical upsampling is skipped, it follows the projection from three-dimensional space to the two-dimensional plane for both point sets. Finally, the proposed attribute upsampling scheme from Sec.~\ref{sec:att_up} follows. The downsampling and upsampling is conducted three times for each point cloud. Thus, the shown results are averaged over all three runs. 
\par As ground truth data is available, we determine the color peak-signal-to-noise ratio (PSNR) as it is known from image processing. Therefore, we first determine the PSNR for each color channel separately and average for Color-PSNR. We measure the color reconstruction PSNR, i.e., the color PSNR is determined solely on the upsampled points.
\par As a second evaluation scheme, we incorporated a histogram comparison as proposed by Viola et al.~\cite{Viola_2020}. As the histograms from reference and upsampled point cloud are compared, it can also be applied if no direct ground truth information is available. The histogram difference of the luminance channel Y is shown in the evaluation with a euclidean distance measure.

\subsection{Influence of Block Parameters}

The block partitioning is a relevant part for the frequency-selective upsampling procedure as it jointly sets the block partitioning for geometry and color upsampling. Hence, the aim is to determine the block parameters such that geometry and attribute quality are as good as possible. Therefore, the influence of block size and size of the newly introduced support area is analyzed in this section. The geometry results are shown in terms of C2C similarity in Fig.~\ref{fig:geo_blkSize} and the color result is depicted in Fig.~\ref{fig:col_blkSize}, respectively. The choice of the parameters affects geometry and attribute upsampling at the same time. Hence, the averaged results for the \textit{3DColorMesh} dataset are depicted for all combinations of core block sizes \(N=2\) (blue), \(N=3\) (red), \(N=4\) (yellow) and \(N=8\) (purple) and support margins relative to block sizes from \(M/N=0\) to \(M/N=1\). The geometry in Fig.~\ref{fig:geo_blkSize} is optimized in terms of C2C similarity as a smooth surface is desired. For block sizes larger than 3, the curves show a slight decrease before the angular similarity increases. For block size $N=2$, the maximum is achieved for a border width of \(M/N=0.25\), i.e. the support margin size is \(M=0.5\). The color differences in Fig.~\ref{fig:col_blkSize} increase with increasing support margins. Hence, color quality is maximized for a model that is as local as possible achieved by a small support margin. In the conjunction of geometry and color quality, the block size is set to \(N=2\) and the support margin is chosen to be \(M=0.5\) for the remainder of this work as geometry is maximized while a good color quality is maintained. 

\subsection{Geometry Results}
Most of the known point cloud upsampling schemes upsample the geometry of a point cloud. We compare the geometry upsampling part of our proposed FSU to FSGU \cite{Heimann_2022_ICIP}. The main difference here is that the block partitioning is conducted differently. In FSU a support area is incorporated whereas FSGU does not take a support area into account. 
In addition, there are two data-driven approaches, namely PU-Net (PU) and EC-Net (EC). PU-Net focuses on adding new points uniformly and as distant from given points as possible whereas EC-Net focuses on upsampling edges properly. For all upsampling techniques, point-to-point (P2P), point-to-plane (P2C) and plane-to-plane (C2C) errors are measured. The results for the 17 point clouds of the \textit{3DColorMesh} dataset are shown in Tab.~\ref{tab:res_geo}. In terms of P2P, the results produced by PU-Net are worst, followed by EC-Net. The best performing approaches are the frequency-model based approaches. Our proposed FSU approach performs slightly worse than FSGU. Due to the incorporated support area in FSU, the upsampled points can be located within the whole block and not just within the range of the original points. Thereby, the distances in between the points increase and thus, also P2P errors may increase. Even though the incorporated support area may not lead to an improvement in the P2P error metric, it leads to an enhanced visual appearance as it is shown in Fig.~\ref{fig:visexample2}. The original of the  \texttt{4armsMonstre} in Fig.~\ref{fig:monstreorig} is upsampled with FSGU in Fig.~\ref{fig:monstrePCS} and our proposed FSU in Fig~\ref{fig:monstreFSU}, respectively. Clear block artifacts can be observed with the upsampling of FSGU. These disappear with FSU such that a better visual quality is achieved. In terms of the metrics, the behavior in terms of P2P metric changes. Here, the FSU results outperform the results generated with FSGU. This shows the improved plane reconstruction in FSU. The results in terms of angular similarity C2C are given in the last columns of Tab.~\ref{tab:res_geo}. As this is a similarity metric, higher values denote better results. Here, the frequency-model based approaches show highest values and thus, are the best upsampling methods for this metric. Hence, FSU outperforms FSGU in terms of P2C error and C2C similarity showing that FSU produces smoother results than FSGU. Thus, the estimation and sampling improved with FSU. The data-driven approaches perform worst.
\par
\begin{figure}
	\resizebox{.9\columnwidth}{!}{%
		\centering
%
\definecolor{mycolor1}{rgb}{0.00000,0.44700,0.74100}%
\definecolor{mycolor2}{rgb}{0.85000,0.32500,0.09800}%
\definecolor{mycolor3}{rgb}{0.92900,0.69400,0.12500}%
\definecolor{mycolor4}{rgb}{0.49400,0.18400,0.55600}%
\definecolor{mycolor5}{rgb}{0.46600,0.67400,0.18800}%
\definecolor{mycolor6}{rgb}{0.30100,0.74500,0.93300}%
\definecolor{mycolor7}{rgb}{0.63500,0.07800,0.18400}%
\begin{tikzpicture}

\begin{axis}[%
	width=.39\textwidth,
	height=.25\textwidth,
at={(2.6in,1.297in)},
scale only axis,
xmin=2,
xmax=4,
ymin=0,
ymax=10,
ylabel style={align=center},
ylabel = {C2C \(\times 10^{-1}\)},
xlabel = {Scaling factor},
xtick distance = 1,
axis background/.style={fill=white},
axis x line*=bottom,
axis y line*=left,
legend columns=4,
legend style={at={(0.5,-0.21)},anchor=north, nodes={scale=0.7, transform shape}}
]

\addplot [color=mycolor2, line width=2pt, mark=cube*]
table[row sep=crcr]{%
	2	4.85\\
	3	4.618\\
	4	4.406\\
};
\addlegendentry{PU (C2C)}

\addplot [color=mycolor3, line width=2pt, mark=cube*]
table[row sep=crcr]{%
	2	3.736\\
	3	3.721\\
	4	3.716\\
};
\addlegendentry{EC (C2C)}

\addplot [color=mycolor4, line width=2pt, mark=cube*]
table[row sep=crcr]{%
	2	7.398\\
	3	6.518\\
	4	6.418\\
};
\addlegendentry{FSGU (C2C)}

\addplot [color=mycolor5, line width=2pt, mark=cube*]
table[row sep=crcr]{%
	2	7.385\\
	3	6.789\\
	4	6.611\\
};
\addlegendentry{FSU (C2C)}


\end{axis}
\end{tikzpicture}%
	}

	\caption{\label{fig:p2p_p2c_scalings} Evolution of the C2C results for the averaged \textit{3DColorMesh} dataset for scaling factors from two to four. Upsampling with PU (orange), EC (yellow), FSGU (purple), and FSU~(green).}
	\vspace{-.3cm}
\end{figure}
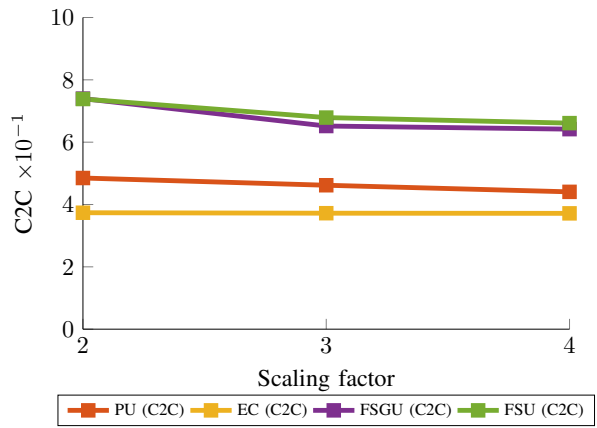

 The behavior in terms of the metrics remain similar for further scaling factors as well. An overview of the averaged result for the dataset from scaling factor two to four is shown in Fig.~\ref{fig:p2p_p2c_scalings} for the angular similarity C2C. The relative behavior as shown in Tab.~\ref{tab:res_geo} of the curve remains constant. FSU is the best performing method in terms of C2C for all scaling factors. As a general trend, the C2C angular similarity metric decreases with increasing scaling factor. 

\begin{table*}
	\caption{\label{tab:res_color}Color results for all point clouds from the \textit{3DColorMesh} dataset for scaling factor of 4 in terms of color PSNR in dB and the histogram distance by Viola et al \cite{Viola_2020}. Best qualities are given in bold. Arrows indicate higher values are better \(\uparrow\) and smaller values are better \(\downarrow\), respectively.}
	\resizebox{\textwidth}{!}{%
		\begin{tabular}{|l|c|c|c|c|c|c||c|c|c|c|c|c|}
			\hline
			&  \multicolumn{6}{c||}{Color PSNR \(\uparrow\)} & \multicolumn{6}{c|}{Histogram distance by Viola et al. \(\times 10\text{e}-2\) \(\downarrow\)} \\
			\hline
			& \multicolumn{2}{c}{3D} & \multicolumn{4}{|c||}{2D} & \multicolumn{2}{c}{3D} & \multicolumn{4}{|c|}{2D} \\
			\hline
			Point Cloud			& LIN3 & NAT3 & LIN2 &  NAT2  &\small FSMMR& FSU& LIN3 & NAT3 & LIN2 &  NAT2  &FSMMR& FSU \\
			\hline
			\hline
			\texttt{4armsMonstre} 	& 22.6	 & 22.6 & 13.0	 & 13.0  & 22.7 & \textbf{25.6} & 3.3 &  3.3 & 31.1 & 31.1   & 1.6 & \textbf{0.7} \\
			\hline
			\texttt{Asterix} 				& 19.9 & 	20.0	 & 7.6	 & 7.6 & \textbf{23.0} & 21.0 & 4.3 & 4.8 &  47.7 & 47.7  &  4.5 & \textbf{3.9}\\
			\hline
			\texttt{CableCar}				& 22.8 & \textbf{23.0 } & 13.1 & 13.1  & 19.0 & 20.7 & \textbf{1.5 }& 1.7 &  20.0 & 20.0  & 2.7   & 2.1\\
			\hline
			\texttt{Dragon}				& 25.8	 & 25.9 & 16.6 & 16.6	 & 23.1 & \textbf{26.2} & 1.8 & 1.9 & 18.9  & 18.9  &  1.8 & \textbf{1.0}\\
			\hline
			\texttt{Duck}						& 12.3  & 	12.4	 & 5.0 &  5.0 & 14.0 & \textbf{15.5} & \textbf{1.8} & 20.4 & 71.2 & 71.3  &  5.7 & 3.0\\
			\hline
			\texttt{GreenDinosaur}	& \textbf{24.6}	 & 24.5	 & 14.0 & 14.0 & 22.2 &	23.4 & 2.0 & 2.2 & 37.6 & 37.6  & 1.7  &\textbf{1.1} \\
			\hline
			\texttt{GreenMonstre}		& 25.0	 & \textbf{25.3}	 &15.2 & 15.2 & 22.2 & 	25.2 & \textbf{1.5} & 1.8 & 16.1  & 16.1  &  3.0 & 2.0 \\
			\hline
			\texttt{Horse}					& 22.3	 & \textbf{22.4}	 & 10.8 & 10.8	 & 17.7 & 19.6 & \textbf{1.7} & 2.1 & 21.0 &  20.9 & 4.3  & 3.2 \\
			\hline
			\texttt{Jaguar}					& 19.8	 & 19.8	 & 13.1 & 13.1 & 23.0& \textbf{25.6} & 3.5 & 3.8 & 10.4  & 10.5  & 4.5 & \textbf{3.2}\\
			\hline
			\texttt{LongDinosaur} 		& 20.1 & 20.2	 &  16.2 & 16.2 & 25.1 & \textbf{27.8 }& 2.9 & 3.0 & 6.7  & 6.7  &  1.9 &\textbf{ 1.4}\\
			\hline
			\texttt{Man}						& 28.7 & \textbf{28.9}	 &  17.6 & 17.6 & 26.8 & 26.5 & 4.1 & 4.2 & 62.6  & 62.6  & 4.1  & \textbf{1.8}\\
			\hline
			\texttt{Mario}					& 22.0	 & \textbf{22.1}& 	14.8 & 14.8  &  22.6 & 20.0 & 3.2 & 4.3 &  6.4 & 6.2  & 3.1 & \textbf{2.5}\\
			\hline
			\texttt{MarioCar}				& 22.7  & 	22.6	 & 14.4 & 14.4  & 22.9  & \textbf{24.4 }& 1.7 & 1.8 & 12.9  & 12.9  & 1.8  & \textbf{1.4} \\
			\hline
			\texttt{PokemonBall}		& 8.7	 & 8.7	 &  7.9 & 7.9	 & 16.8 & \textbf{18.6}& 25.6 & 25.4 & 41.5 & 41.5  &   8.9 &\textbf{ 5.0} \\
			\hline
			\texttt{Rabbit}					& 20.0	 & 20.0 & 	10.1 & 10.1 & 21.1 & \textbf{23.4} & \textbf{2.8} & 3.2 & 20.2  & 20.2  &  5.0 & 3.4 \\
			\hline
			\texttt{RedHorse}			& 21.4 & 	\textbf{21.5} & 13.2 & 13.2 & 18.9 & 21.0& 1.6 & 1.7 & 14.5  & 14.5 & 2.3 & \textbf{1.3} \\
			\hline
			\texttt{Statue}					& 22.9 &  22.9 & 	14.1	 & 14.1	 & 21.8 & \textbf{24.2}& 2.6 & 2.7 & 21.3  & 21.3   & 1.5 & \textbf{0.7}\\
			\hline
			\hline
			Average 			& 21.3 &  21.3 & 	12.7	 & 12.7	 & 21.0 & \textbf{22.9} & 4.8 & 5.2 & 27.1 & 27.1   & 3.4  &  \textbf{2.2} \\
			\hline
		\end{tabular}
	}
	\vspace{-.3cm}
\end{table*}
\vspace{-.2cm}

\subsection{Color Results}
\label{sec:col_results}
In a second evaluation, we analyze the quality of the color attribute of the point clouds. Our proposed upsampling scheme is denoted as FSU. We compare FSU to the Frequency-Selective Mesh-to-Mesh Resampling (FSMMR) as proposed in \cite{Heimann_2021_MMSP}, linear interpolation on block-level (LIN2) and on the whole point cloud (LIN3). Furthermore, natural neighbor interpolation is also appplied on both, block level (NAT2) and on the whole point cloud~(NAT3). 

\begin{figure*}
		\vspace{-.15cm}
	\begin{subfigure}[t]{.33\textwidth}
		\centering
			\includegraphics[scale=.3, trim=410 135 420 110 mm, clip=true]{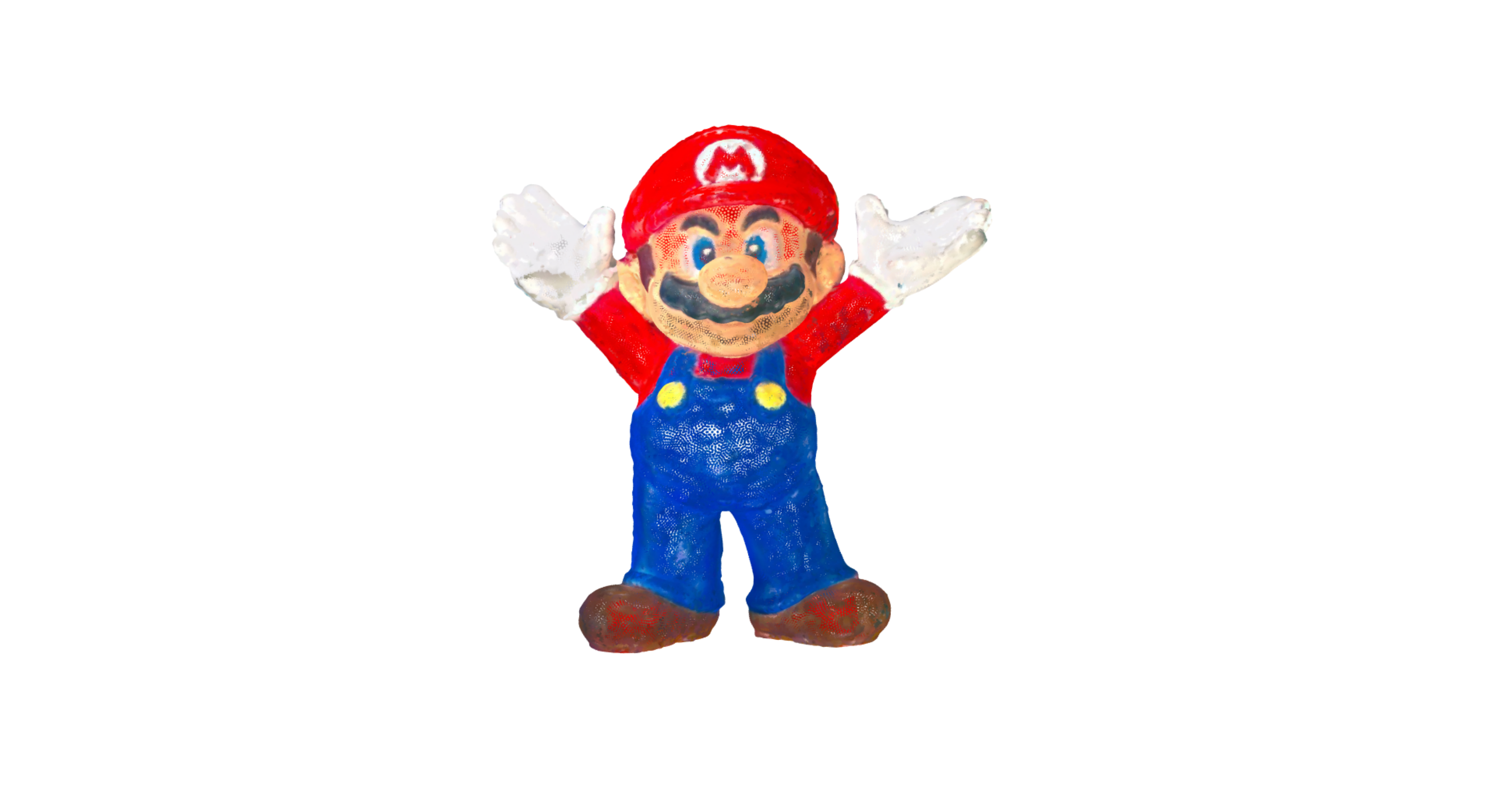}
		\caption{\label{fig:marioorig}Original.}
	\end{subfigure}
	\begin{subfigure}[t]{.33\textwidth}
		\centering
		\includegraphics[scale=.3, trim=410 135 420 110 mm, clip=true]{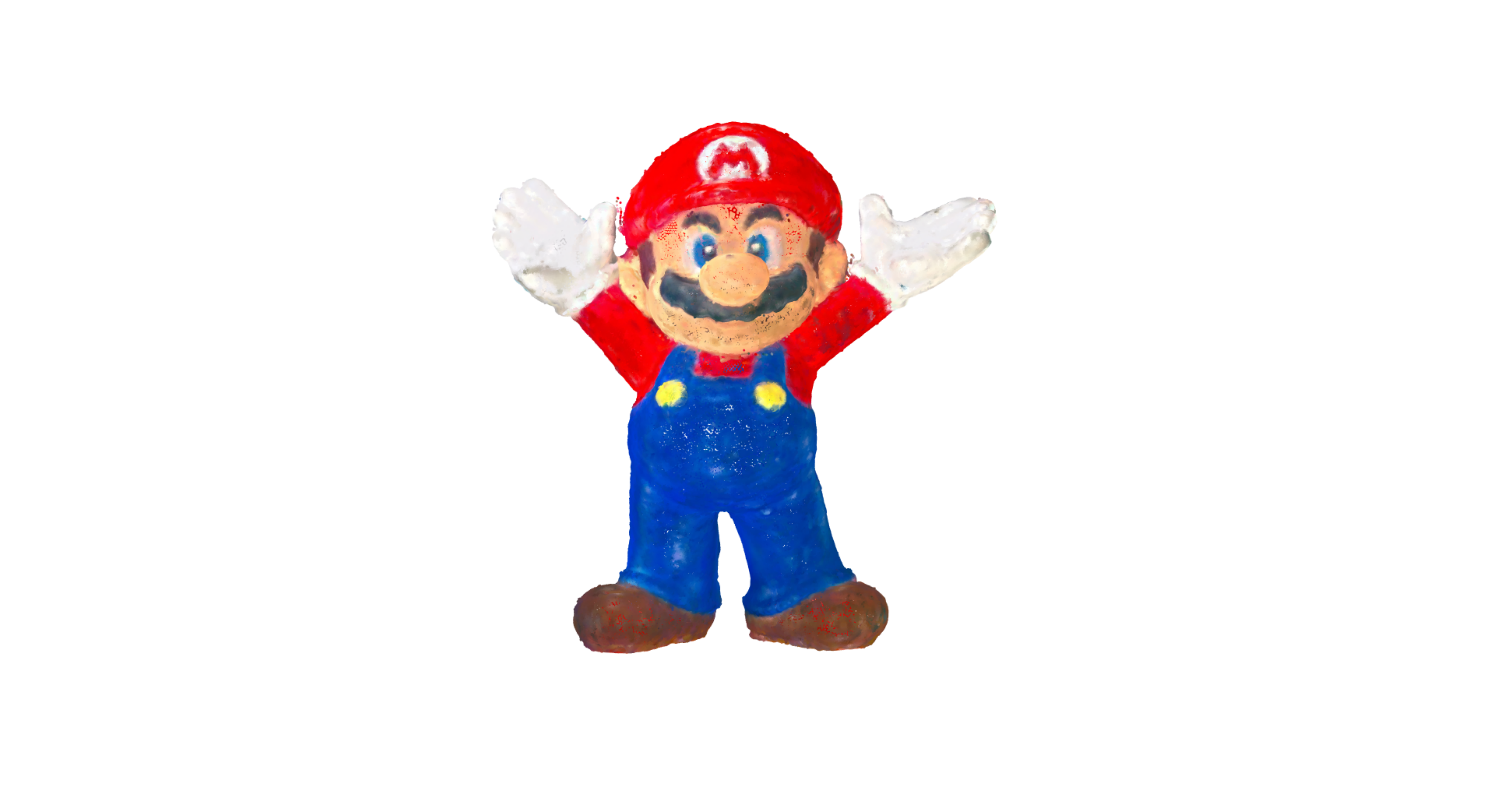}
		\caption{\label{fig:mariopcs} FSGU+FSMMR.}
	\end{subfigure}
	\begin{subfigure}[t]{.33\textwidth}
		\centering
		\includegraphics[scale=.3, trim=410 135 420 110 mm, clip=true]{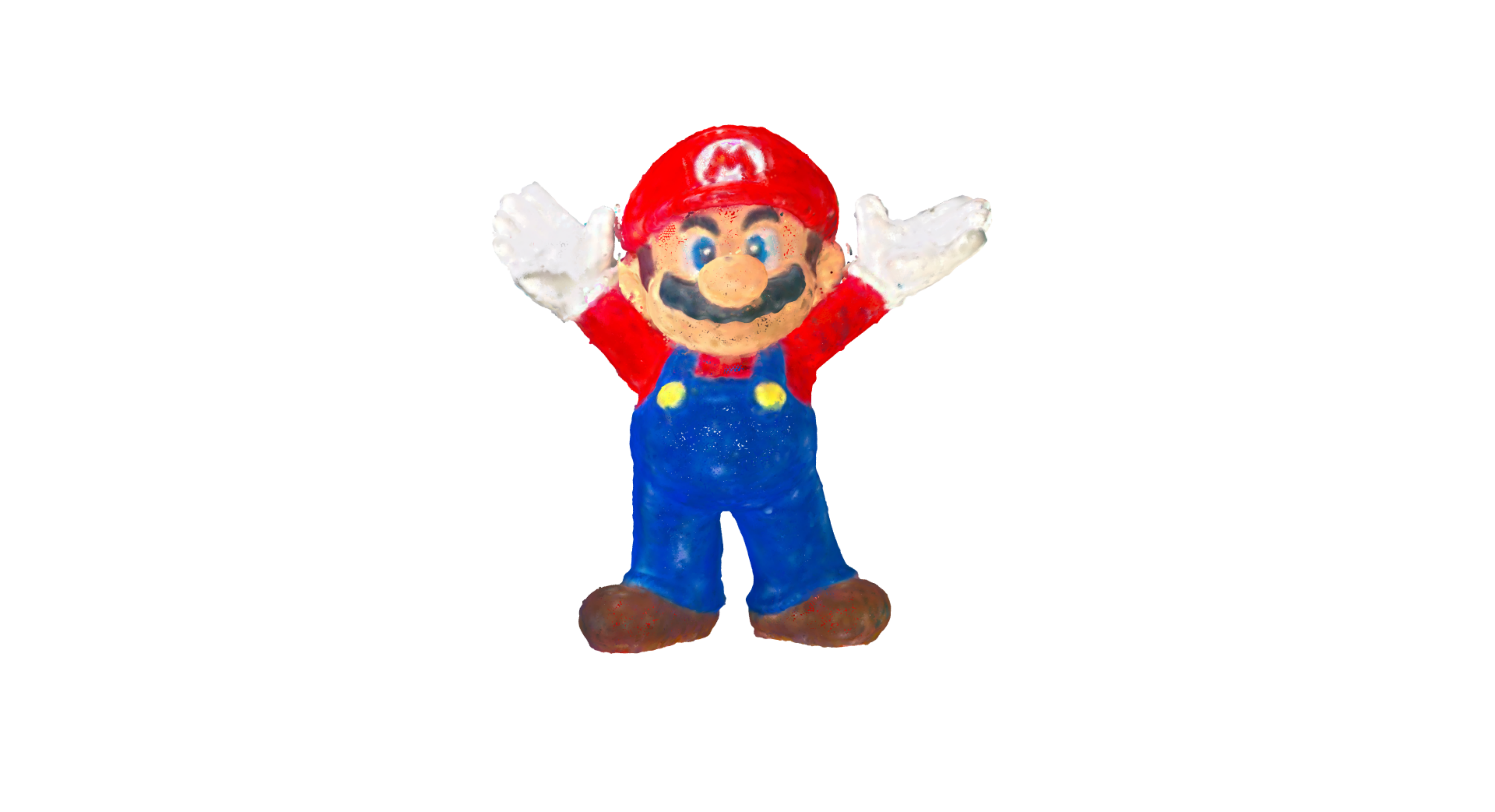}
		\caption{\label{fig:mariofsu} FSU (Proposed).}
	\end{subfigure}
	\caption{\label{fig:visexample} \texttt{Mario} point cloud. Upsampling factor is 4. }
	\vspace{-.3cm}
\end{figure*}

The results for the metrics introduced in Sec.~\ref{sec:metric_att} are given in Tab.~\ref{tab:res_color}. The upsampling techniques are evaluated for all 17 point clouds from the \textit{3DColorMesh} dataset. The color PSNR is shown in the first six columns. The results of the histogram based evaluation of Viola et al.~\cite{Viola_2020}~is depicted in the last six columns of the table. As it is shown in terms of PSNR, the proposed FSU performs best on average with a gain of $~1.9$~dB to FSMMR. Especially for the \texttt{Dragon} point cloud, gains of up to $3.1$~dB are achieved. The results for our proposed approach FSU show that it is advantageous to include a support area and thereby, take more neighborhood information into account for the model estimation. Severe degradations of the point clouds can be observed for the interpolation approaches as they are based on triangulations. Hence, color reconstruction may not be possible in concave areas where an extrapolation is necessary. Thus, not all color information can be retrieved and thus, color PSNR degrades. A similar behavior can be observed for the histogram difference by Viola et al.~\cite{Viola_2020} in the last six columns of the table. As it is a distance-based measure, smaller values indicate better results. Severe degradations can be observed for the interpolation-based approaches especially for the \texttt{Duck}, \texttt{Man}, and \texttt{PokemonBall} point clouds. As our model-based FSU can retrieve all color information independently of whether inter- or extrapolation is required the histogram distances are much smaller. On average, FSU performs best in terms of color PSNR and in terms of histogram distances.  

\subsection{Joint Results}
\begin{figure*}
	\vspace{-.15cm}
	\begin{subfigure}[t]{.33\textwidth}
		\centering
		\includegraphics[scale=.2, trim=430 105 400 50 mm, clip=true]{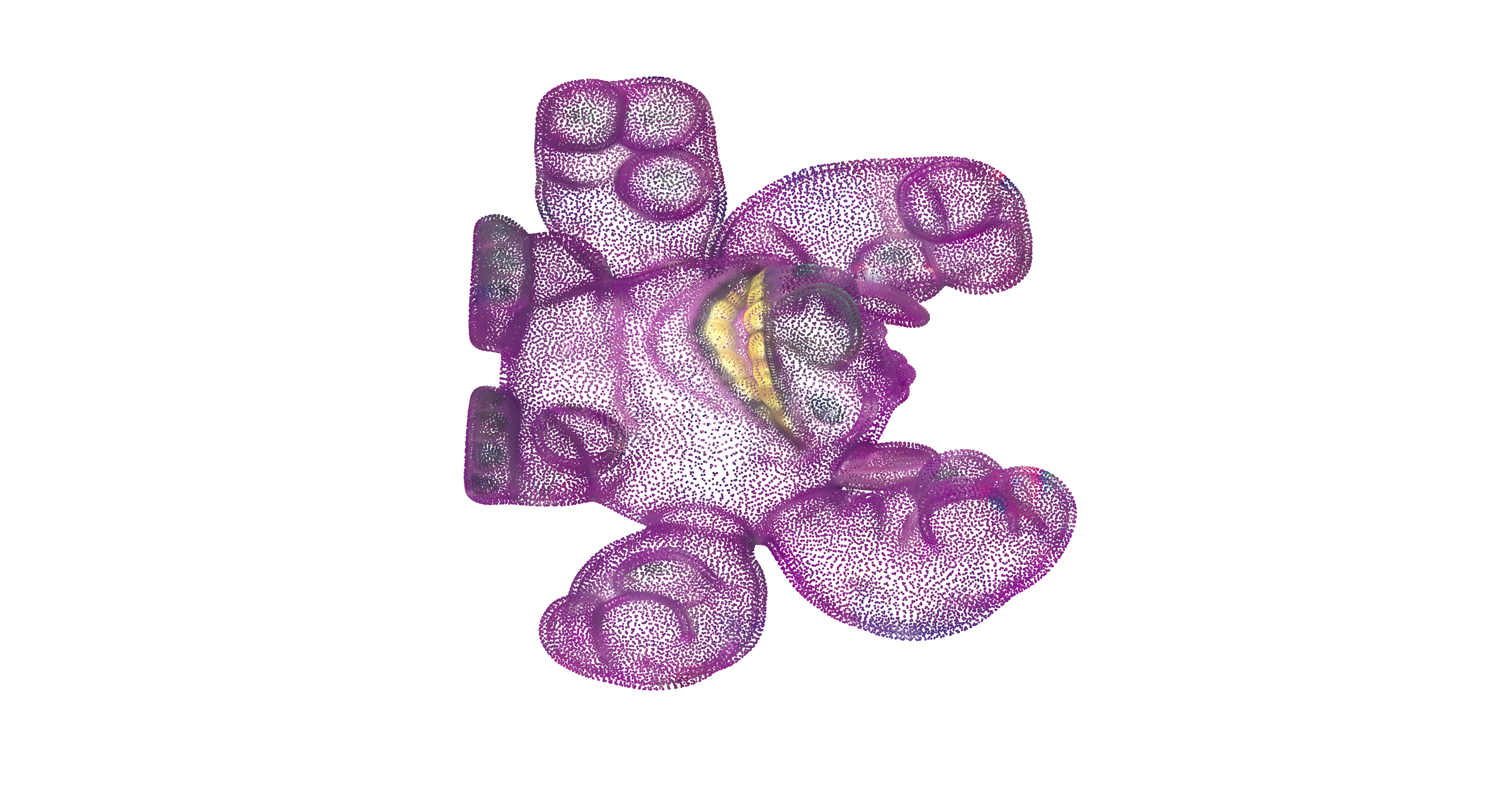}
		\caption{\label{fig:monstreorig}Original.}
	\end{subfigure}
	\begin{subfigure}[t]{.33\textwidth}
		\centering
		\includegraphics[scale=.2, trim=430 105 400 50 mm, clip=true, angle=180, origin=c]{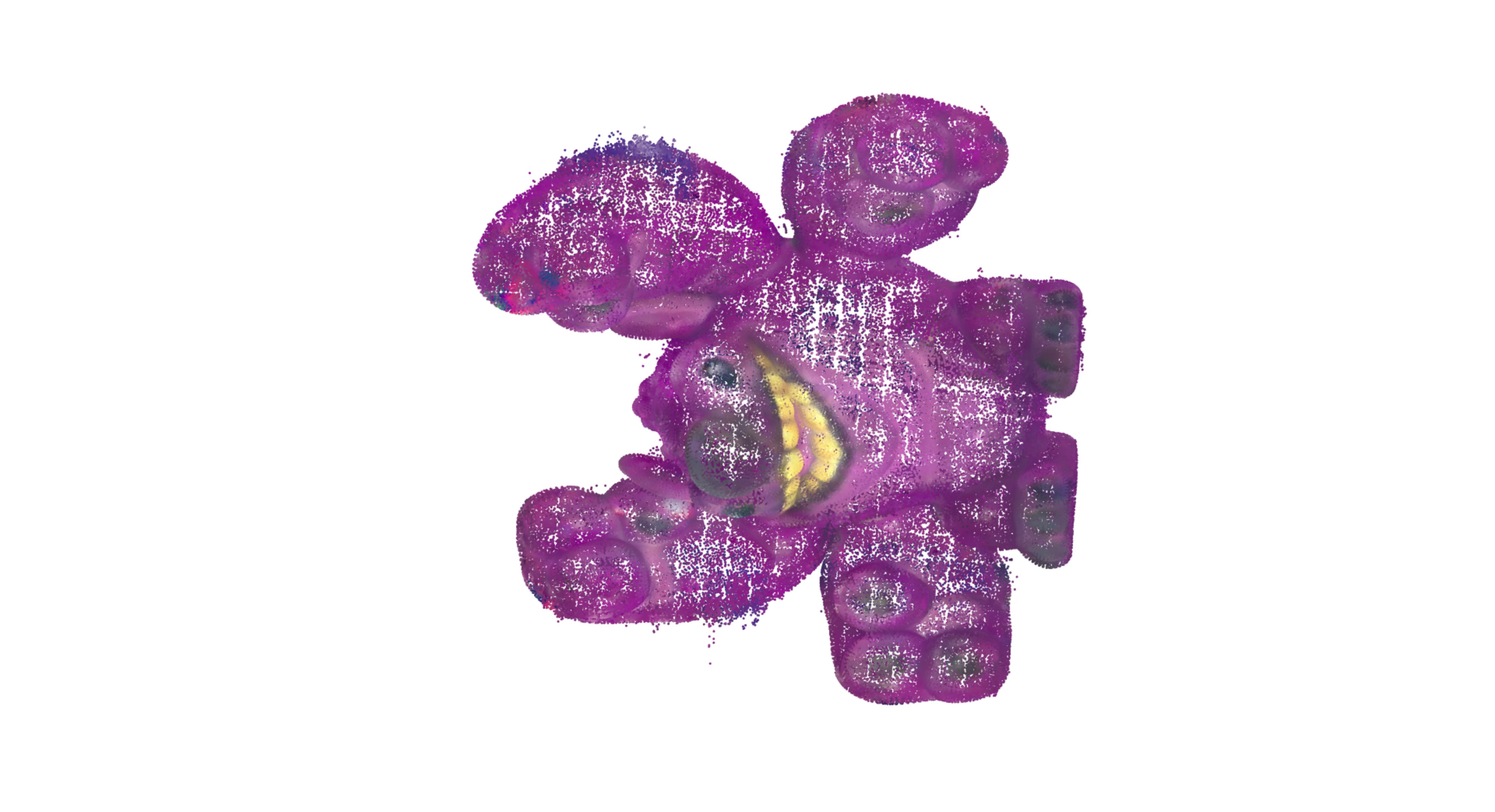}
		\caption{\label{fig:monstrePCS} FSGU + FSMMR.}
	\end{subfigure}
	\begin{subfigure}[t]{.33\textwidth}
		\centering
		\includegraphics[scale=.2, trim=430 105 400 50 mm, clip=true]{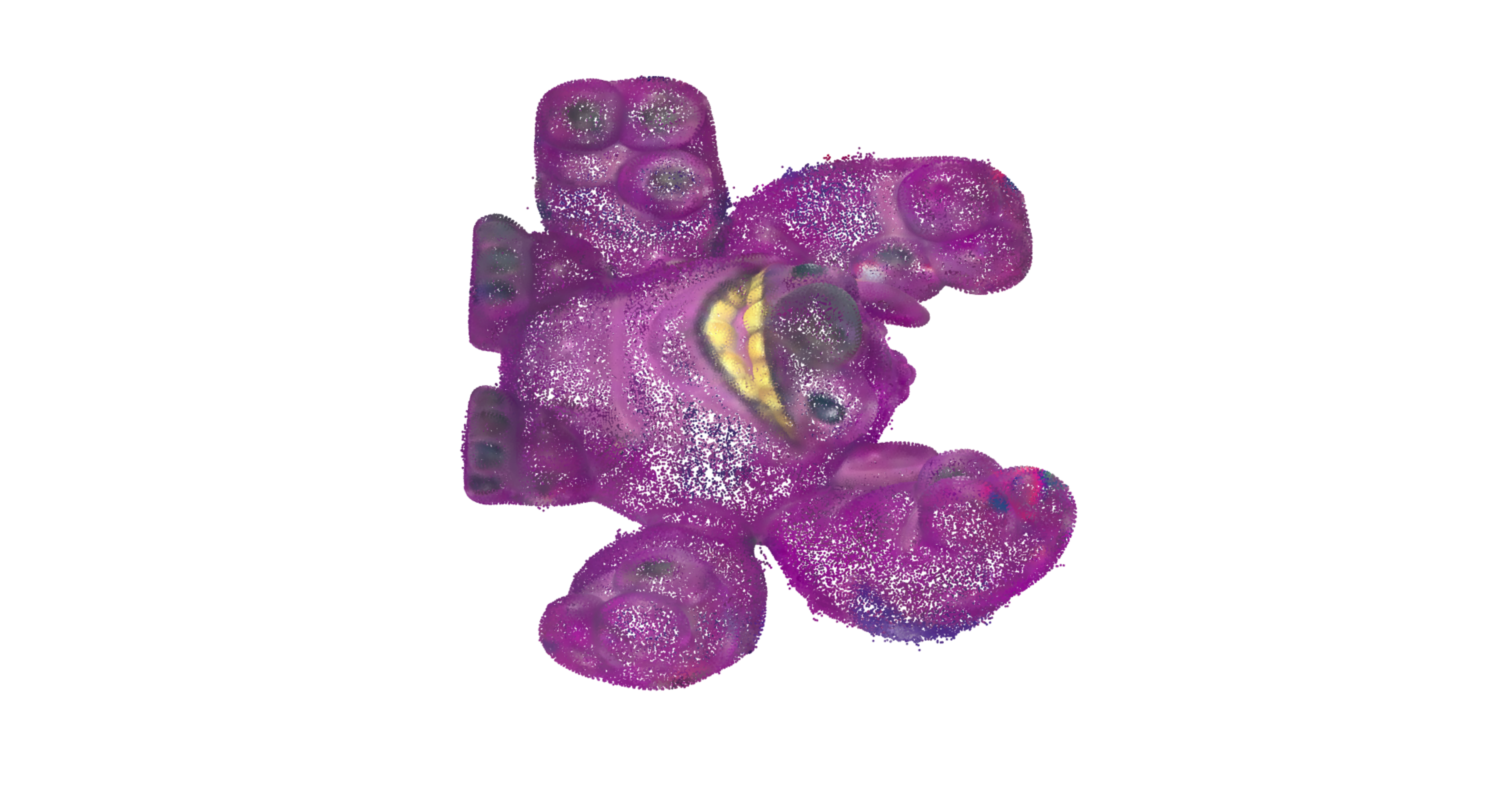}
		\caption{\label{fig:monstreFSU}FSU (Proposed).}
	\end{subfigure}
	\caption{\label{fig:visexample2} \texttt{4armsMonstre} point cloud. Upsampling factor is 4.}
		\vspace{-.5cm}
\end{figure*}

Our proposed FSU upsamples geometry and attribute jointly. Some visual results are given in Figs.~\ref{fig:visexample}~and~\ref{fig:visexample2}. The original point cloud with original resolution is given in Subfigs. (a). Subfigs. (b) depict the results if FSGU~\cite{Heimann_2022_ICIP} and FSMMR~\cite{Heimann_2021_MMSP} , are combined. Subfigs. (c) depict the proposed FSU. Clear improvements can be observed for our proposed FSU. Especially the \texttt{4armsMonstre} shows clear blocking artifacts in Fig.~\ref{fig:monstrePCS}. The newly proposed FSU as given in Fig.~\ref{fig:monstreFSU} improves these artifacts notably and appears to be much smoother. This is mainly due to the added support area that is used during the model generation. 

\section{Conclusion}
\label{sec:con}
In this work, we presented a joint upsampling scheme for geometry and attribute of point clouds. We therefore incorporate frequency-selective models. The models are estimated locally on block level. In the block partitioning an overlapping support area is included that incorporates neighborhood information into the block estimation process. For attribute upsampling, information from the geometry upsampling step is exploited. Our proposed Frequency-Selective Upsampling (FSU) improves point cloud upsampling in both, geometry and color quality. FSU shows best results in terms of point-to-plane error and plane-to-plane angular similarity. Furthermore, the color upsampling quality is on average improved by \(1.9\)~dB in terms of color PSNR. Also the visual appearance of the upsampled point cloud is improved notably as the included support area reduces block artifacts clearly. 
\balance

\bibliographystyle{IEEEtran}
\small{
\bibliography{bib_4tmm2022}
}

\end{document}